\documentclass[preprint,3p,times]{elsarticle}

\usepackage{amssymb}
\usepackage{amsthm}
\usepackage{gensymb}
\usepackage{lineno}
\modulolinenumbers[5]
\usepackage{caption}
\usepackage[fleqn]{amsmath}
\usepackage{multirow}

\usepackage{hyperref}
\biboptions{sort&compress}
\hypersetup{colorlinks = true, allcolors = blue, pdfauthor=author}
\usepackage[nameinlink,capitalise]{cleveref}

\begin{document}

\begin{frontmatter}



\title{Characterising the take-off dynamics and energy efficiency in spring-driven jumping robots}


\author[inst1]{John Lo \corref{cor1}}
\cortext[cor1]{Corresponding author.}
\ead{kwokcheungjohn.lo@manchester.ac.uk}

\author[inst1,inst2]{Ben Parslew}
\affiliation[inst1]{organization={Department of Fluid and Environment},
            addressline={The University of Manchester}, 
            city={Manchester},
            postcode={M13 9PL}, 
            country={United Kingdom}}

\affiliation[inst2]{organization={International School of Engineering, Faculty of Engineering
},
            addressline={Chulalongkorn University}, 
            city={Bangkok},
            postcode={10330}, 
            country={Thailand}}

\begin{abstract}
Previous design methodologies for spring-driven jumping robots focused on jump height optimization for specific tasks. In doing so, numerous designs have been proposed including using nonlinear spring-linkages to increase the elastic energy storage and jump height. However, these systems can never achieve their theoretical maximum jump height due to taking off before the spring energy is fully released, resulting in an incomplete transfer of stored elastic energy to gravitational potential energy. This paper presents low-order models aimed at characterising the energy conversion during the acceleration phase of jumping. It also proposes practical solutions for increasing the energy efficiency of jumping robots. A dynamic analysis is conducted on a multibody system comprised of rotational links, which is experimentally validated using a physical demonstrator. The analysis reveals that inefficient energy conversion is attributed to inertial effects caused by rotational and unsprung masses. Since these masses cannot be entirely eliminated from a physical linkage, a practical approach to improving energy efficiency involves structural redesign to reduce structural mass and moments of inertia while maintaining compliance with structural strength and stiffness requirements.
\end{abstract}


\end{frontmatter}


\section{Introduction}
\label{sec:intro}
\begin{figure}[t]
\centering
\includegraphics[scale=1]{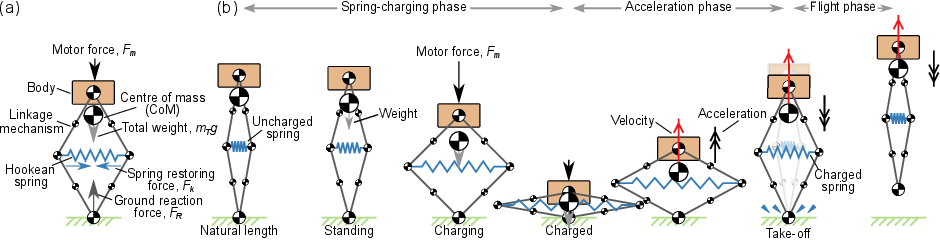}
\caption{An example of (a) a nonlinear spring-driven jumping robot with a rotational linkage mechanism that exhibits a nonlinear force-length relationship (e.g. \citep{JPL_Hale_2000}) and (b) its entire jumping process.}
\label{fig:intro}
\end{figure}

Spring-driven jumping robots use a motor to store elastic energy in a spring, and then release this energy to propel the system \citep{NatureJ_Hawkes_2022,Ilton_2018} (\cref{fig:intro}).   For an equivalent peak force and characteristic length, springs can generate greater output power and do more work over the system stroke length than a standalone motor \citep{NatureJ_Hawkes_2022,Ilton_2018}. This technique is known as \textit{power amplification} and has been widely used in jumping animals, such as fleas \citep{Flea_Sutton_2011,Flea_BennetClark_1967} and locusts \citep{Locust_Wan_2016,Locust_Burrows_2012,Locust_BennetClark_1975}. Generally, the primary objective in designing spring-driven jumping robots is to achieve high \textit{mechanical-kinetic energy efficiency} \citep{JReview_Zhang_2017,JReview_Zhang_2020,JReview_Ribak_2020,JReview_Mo_2020}. This involves having high \textit{mechanical-elastic} efficiency in transferring energy from a stored onboard source via a motor to a spring during the spring-charging phase. It also requires high \textit{elastic-kinetic} efficiency when transferring the spring elastic energy to kinetic energy as the robot accelerates during the acceleration phase. This paper focuses on the latter of these two phases.

The elastic-kinetic energy conversion efficiency in spring-driven jumping robots was first explored in detail by \citep{JPL_Fiorini_1999} while designing a prismatic jumper for planetary exploration propelled by a linear spring (obeying Hooke’s law). A critical issue cited was that during the acceleration phase the robot foot disengaged from the ground before the leg was fully extended \citep{JPL_Hale_2000}. This resulted in the robot taking off prematurely with only 20\% of the stored elastic potential energy being converted to kinetic energy (termed \textit{premature take-off}). Hence, the robot did not reach the jump height predicted by the models, which assumed 100\% elastic-kinetic conversion efficiency. The group attributed this premature take-off to spring surge \citep{JPL_Hale_2000} – a dynamic effect where the spring natural frequency is excited by the acceleration of the robot \citep{SpringSurge_1996} – but no specific evidence was provided of this phenomenon. The analysis in \cref{sec:baton} of this work will directly address the problem of premature take-off, and provide evidence that it is caused by the centripetal force acting on the rotational bodies of the system and not by spring surge. 

A follow-on study proposed an alternative closed-chain linkage with a linear spring (hereafter referred to as \textit{spring-linkage}, \cref{fig:intro}) to mitigate premature take-off in spring-driven systems \citep{JPL_Hale_2000}. The closed-chain linkage exhibited a nonlinear force-displacement relationship when loaded and had higher mechanical-elastic energy conversion efficiency compared to a linear spring \citep{LO_Statics_2023,LO_2021}. The authors attributed the improved jump height in \citep{JPL_Hale_2000} to the nonlinearity of the force profile that allowed more energy to be transferred at the early stage of the jump and less at the end. The authors argued that this energy release profile would reduce the impact of premature take-off on the elastic-kinetic conversion efficiency. However, they did not consider that the greater jump height achieved by the spring-linkage may simply be due to the greater elastic energy storage compared to the previous linear spring system. 

The significance of the spring-linkage design proposed in \citep{JPL_Hale_2000} is that it sparked the development of a whole class of similar closed-chained spring-linkage mechanisms \cite{Jumproach_Jung_2016,Microrobot_Bergbreiter_2007,FleaJ_Koh_2013,ScoutJ_Zhao_2009,ESJ_Bai_2018,Multimobat_Woodward_2014,PEAnSEA_Hong_2020,Controllable_miniatureJ_Zhao_2010,SurveillanceJ_Song_2009,SpaceJ_Wang_2022,Flysquirrel_Zhao_2021,SphereJ_Chang_2022}; this topology has since become the most prevalent in jumping robot designs. Several mechanisms have demonstrated improvements in the mechanical-elastic energy transfer efficiency over the original design \citep{JPL_Hale_2000} by using different types and arrangements of elastic elements \citep{LO_2021,LO_Statics_2023,FlapJ_Truong_2019,Jumproach_Jung_2019,MSU_Zhao_2013,TAUB_Zaitsev_2015,Jollbot_Armour_2008,NatureJ_Hawkes_2022,Springtail_Ma_2021,TSJ_Zhang_2018}. However, the spring arrangement does not influence the premature take-off: all previous spring-driven robots have had premature take-offs. This is important as premature take-off limits the overall mechanical-kinetic energy conversion. The contribution of this paper is to identify and describe the causes of premature take-off and sources of energy inefficiency in spring-driven jumping robots. This will be tackled using simple conceptual models to provide a transparent analysis. The findings will then be applied to more complex systems to provide practical engineering solutions to increase energy conversion efficiency.

\section{Background theory of take-offs in spring-driven jumping systems}
\label{sec:background}
This work classifies the take-offs of a spring-driven system into three categories: (1) \textit{idealised take-off}, where at take off the spring is at its natural length and all stored elastic potential energy has been fully released; (2) \textit{premature take-off}, where at take off the spring length is less than the natural length and so still stores elastic potential energy;  (3) \textit{delayed take-off}, where at take off the spring length is greater than its natural length and so still stores elastic potential energy. Premature and delayed take-offs are undesirable as the residual elastic energy in the spring lowers the elastic-kinetic energy conversion efficiency and overall mechanical-kinetic conversion efficiency. 

This section will illustrate the cause of the three take-off categories in spring-driven jumpers using two simple dynamic models: a prismatic multibody system driven by a translational spring (\cref{sec:prismatic}), and a rotational rigid body driven by a rotational spring (\cref{sec:baton}). These are conceptually the simplest models explaining the jumping dynamics that still apply to more complicated mechanical systems.   

\subsection{Prismatic linkage jumping systems}
\label{sec:prismatic}

\begin{figure}[hb]
\centering
\includegraphics[scale=1]{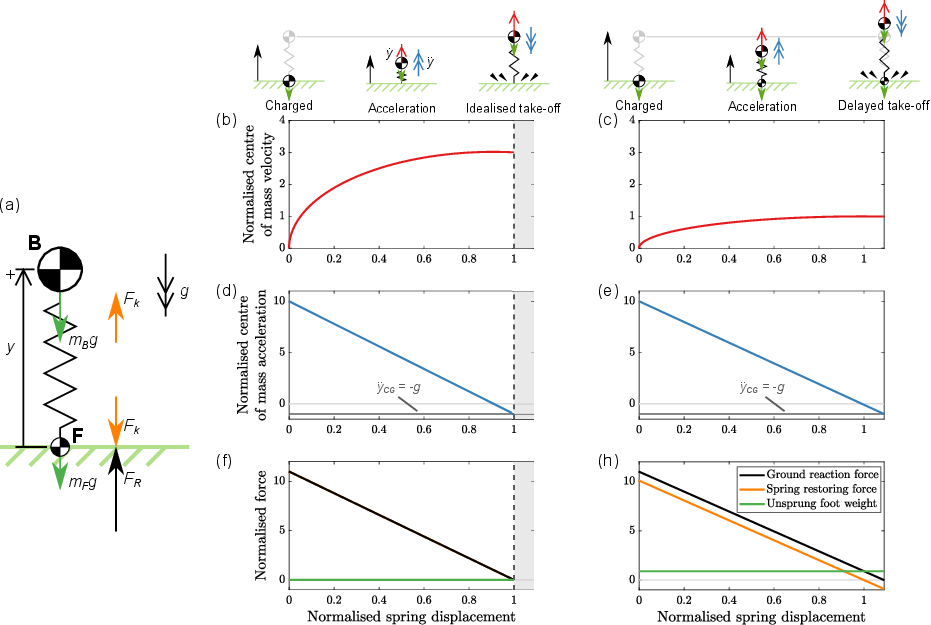}
\caption{(a) A prismatic jumping model driven by a massless linear spring in the standing posture. (b-c) The velocity and (d-e) acceleration of the centre of mass and (f-g) ground reaction force of the prismatic model (a) with two different body mass fractions: 1 (b,d,f) and 0.1 (c,e,h), both at a given force-to-weight ratio of 10. Without the unsprung mass (b) will take off as the spring returns to its natural length and, thus, achieves an idealised take-off. With an unsprung mass at the foot (c) undergoes a delayed take-off. Note that the greyed area is the \textit{inflight phase} after the model takes off, which is out of the scope of this study. The spring displacement is normalised by the natural spring length as $y/d$; the velocity of the centre of mass is normalised by the gravitational acceleration and characteristic length as, $\dot{y}_{CG}/\sqrt{gd}$; the acceleration of the centre of mass is $\ddot{y}_{CG}/g$. The force is normalised by the peak spring charging force, $F/F_{max}$.}
\label{fig:prismatics}
\end{figure}

The mass spring system depicted in \cref{fig:prismatics} is the simplest dynamic system that can be used to illustrate delayed take-off. The jumping dynamics of a linear (Hookean) spring-driven prismatic jumping system have been previously examined in \citep{Jumpingtakeoff_Parslew_2018}; this work extends the original analysis by studying the influence of the unsprung mass  on the dynamics and the energy efficiency. The prismatic jumping model is modelled as a lumped sprung \textit{body} mass (\cref{fig:prismatics}a, point B) connected to an unsprung \textit{foot} mass (point F) via a massless translational linear spring (\cref{fig:prismatics}a). The spring has a stiffness, $k$, and a natural length, $d$, which is also defined as the characteristic length of this model. As the mechanical interface between the model and the ground, take-off is defined as the time when the foot loses contact with the ground. The system kinematics are captured through the body displacement, $y$, and the displacement of the system centre of mass, $y_{CG}$; note that $y_{CG}=[m_B/(m_B+m_F)]y$, where $m_B$ is the mass of the body and $m_F$ is the mass of the foot. The \textit{body-mass fraction} is the ratio of the body mass to the total system mass, $m_B/m_T$, where $m_T=m_B+m_F$. When the body mass fraction equals zero, all the system mass is located at the unsprung foot and there is zero sprung mass. Conversely, if the body mass fraction equals one, the system has zero foot mass and is the same as the ideal prismatic model presented in \citep{Jumpingtakeoff_Parslew_2018}. 

The jumping process begins from the standing posture where the spring is already displaced from its natural length due to the body weight, $m_Bg$ (\cref{fig:prismatics}a). The motor force, $F_m(y)$,  does work on the spring until the force reaches the peak force, $F_{max}$ and the spring is defined as being at the \textit{charged} state, where $y=y_{CG}=0$. The \textit{force-to-weight ratio} is defined as the ratio of the peak force to the total system weight, $\alpha=F_{max}/m_Tg$.  During this process, the work done by the motor force is stored as elastic potential energy in the linear spring as, $EPE=\frac{1}{2}kd^2=\frac{1}{2}(F_{max}+m_Bg)d$. Note that the body weight contributes to the stored elastic energy, but the foot weight does not. When the motor force is removed (which could be achieved mechanically by using a latch) at the beginning of the acceleration phase, the elastic potential energy is transferred to kinetic energy and the mass accelerates in the $+y$ direction. The equation of motion of the body as it accelerates is,

\begin{equation}
    F_k(y)-m_Bg=k(d-y)-m_Bg=m_B\ddot{y},\label{eq_pris_EOM}
\end{equation}
where $F_k$ is the spring restoring force. The solutions of \cref{eq_pris_EOM} are $y=\frac{kd-m_Bg}{k}[1-\cos{(\sqrt{\frac{k}{m_B}}t)}]$ for the body displacement, $\dot{y}=\frac{kd-m_Bg}{\sqrt{km_B}}\sin{(\sqrt{\frac{k}{m_B}}t)}$ for body velocity and $\ddot{y}=\frac{kd-m_Bg}{m_B}\cos{(\sqrt{\frac{k}{m_B}}t)}$ for body acceleration. As shown in \cref{fig:prismatics}a, the ground reaction force acting on the foot is the sum of the spring restoring force and the unsprung weight,
\begin{equation}
    F_R=F_k(y)+m_Fg=k(d-y)+m_Fg=m_T\ddot{y}_{CG}+m_Tg,\label{eq_pris_FR}
\end{equation}
where $\ddot{y}_{CG}=\frac{m_B}{m_T}\ddot{y}$, is the acceleration of the centre of mass of the system. Take-off occurs when the ground reaction force equals zero. Hence, from \cref{eq_pris_FR}, the body displacement at take-off, $y_{to}$, can be expressed as,
\begin{equation}
    y_{to}=d+\frac{m_Fg}{k};\label{eq_pris_yto}
\end{equation}
the take-off acceleration of the centre of mass is, $\ddot{y}_{CG,to}=-g$ and; the take-off acceleration of the body is, $\ddot{y}_{to}=-\frac{m_T}{m_B}g$. By substituting this to the solution for the acceleration of the body, the time of the take-off can be expressed as, $t_{to}=\sqrt{\frac{m_B}{k}}\cos^{-1}{(\frac{-m_Tg}{m_B-kd})}$. Subsequently, by substituting the take-off time into the solution for the velocity of the body,  the take-off velocity of the body is, $\dot{y}_{to}=\sqrt{\frac{kd(kd-2m_Bg)-m_Fg^2(2m_B-m_F)}{km_B}}$; the take-off velocity of the centre of mass is 
\begin{equation}
    \dot{y}_{CG,to}=\frac{m_B}{m_T}\dot{y}_{to}=\frac{1}{m_T}\sqrt{\frac{m_B}{k}[kd(kd-2m_Bg)-m_Fg^2(2m_B-m_F)]}.\label{eq_pris_vyCGto}
\end{equation}

The elastic-kinetic energy conversion efficiency, $\varepsilon$, is explicitly defined here as the ratio of the vertical take-off kinetic energy of the centre of mass, $KE_{y,CG,to}$, to the initial elastic potential energy stored in the spring, $EPE_{ini}$, as,
\begin{equation}
    \varepsilon=\frac{KE_{y,CG,to}}{EPE_{ini}}=\frac{m_T(\dot{y}_{CG,to})^2}{kd^2}=\frac{m_B}{m_Tk^2d^2}[kd(kd-2m_Bg)-m_Fg^2(2m_B-m_F)].\label{eq_pris_eff}
\end{equation}

For the given prismatic model, the elastic-kinetic energy efficiency is maximised when the unsprung mass is zero. However, it is noteworthy that in any spring-driven jumping system operating under nonzero gravitational acceleration, the elastic-kinetic energy conversion approaches to, but never equals, one. This is because part of the energy must be converted into gravitational potential gained by the body before take-off.  

\cref{fig:prismatics} shows examples of the jumping dynamics during the acceleration phase of the prismatic model. The results are presented for two body mass fractions: 1 (\cref{fig:prismatics}b,d,f) and 0.1 (\cref{fig:prismatics}c,e,g), for a given characteristic length and force-to-weight ratio. As the spring is released, it applies a positive force to the system mass, causing the body mass to displace and the velocity of the centre of mass to increase (\cref{fig:prismatics}b,c). Acceleration of the centre of mass decreases proportionally with the spring displacement, as the spring is Hookean. Take-off occurs when the ground reaction force equals zero and the acceleration of the centre of mass equals $-g$ (\cref{fig:prismatics}d,e).

With zero unsprung mass (\cref{fig:prismatics}b,d,f or in \citep{Jumpingtakeoff_Parslew_2018}) or zero gravitational acceleration ($g=0$), an idealised take-off is realised as given from \cref{eq_pris_yto}. Here, the system takes off when the spring length is equal to its natural length. The stored elastic potential energy is fully released from the spring before take-off; part of it is converted into the kinetic energy of the centre of mass, while the remainder is transformed into the gravitational potential energy of the centre of mass.

With nonzero unsprung mass (e.g. \cref{fig:prismatics}c,e,g), the system undergoes delayed take-off as the spring is extended beyond its natural length at take-off. At take-off the spring applies a force equal to but opposite the unsprung weight (\cref{fig:prismatics}g). As a result, the system takes off with nonzero residual elastic energy remaining in the spring, leading to a reduced centre of mass velocity compared to the zero unsprung mass model. Delayed take-off is undesirable as it reduces the elastic-kinetic energy conversion efficiency and take-off velocity of the centre of mass. It is exacerbated by the increase in unsprung mass as body mass fraction decreases (\cref{eq_pris_yto}). After take-off this residual elastic energy is exchanged with the kinetic energy of the masses, causing them to oscillate relative to the system’s centre of mass, although this does not influence the jump height of the centre of mass.

In practice, all physical prismatic systems must have a nonzero unsprung interface (e.g. foot) between the mechanism and the ground. Therefore, it can be concluded that no physical prismatic spring-driven systems can achieve an ideal take-off.

Finally, it should be highlighted that for a premature take-off to occur, the ground reaction force is zero while the spring length is less than its natural length. However, this condition cannot be met in this linear spring model assumed here. \cref{sec:baton} will show that a spring-driven system with a rotational body can undergo premature take-off.  

\subsection{Rotational linkage jumping systems}
\label{sec:baton}

\begin{figure}[ht]
\centering
\includegraphics[scale=1]{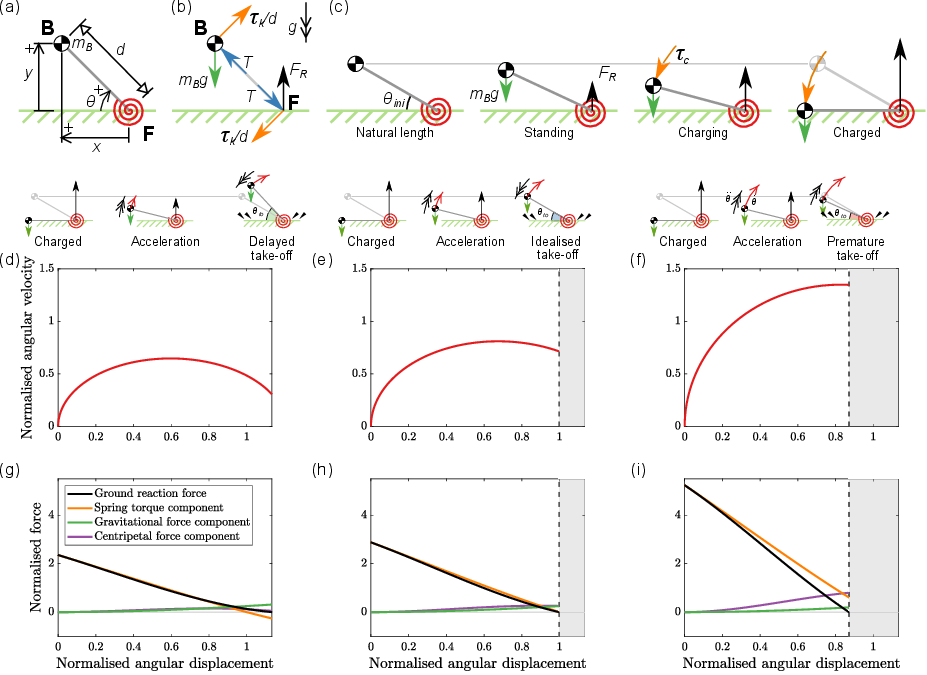}
\caption{(a) An inverted baton model driven by a massless rotational spring, and (b) its free-body diagram during the acceleration phase. (c) The spring-charging phase of the model with a natural angle of 30$\degree$. (d,e,f) The angular velocity and (g,h,i) the ground reaction force of the rotational model as a function of normalised angular displacement. The model in (d-i) has three different normalised rotational spring stiffness, $k_r/mgd$: (d,g) 4.5, (e,h) 5.5, (f,i) 10, all at a given natural angle of 30$\degree$. Note that the given natural spring angle is selected arbitrarily. For a given normalised rotational spring stiffness, an increase in natural spring angle exacerbates the premature take-off. This is because an increase in the natural spring angle increases the spring restoring torque, which increases the centripetal force and subsequently causes more premature take-off. The greyed areas in (e-f) and (h-i) are the \textit{inflight phase} after the model takes off, which is out of the scope of this study. The angular displacement is normalised by the natural spring angle, $\theta/\theta_{ini}$ and the angular velocity is normalised by the gravitational acceleration and the characteristic length, $\dot{\theta}=\sqrt{d/g}$; the force is normalised by the weight of the rotational system, $F/m_Bg$.}
\label{fig:baton}
\end{figure}

A single-degree-of-freedom linkage with a rotational joint can demonstrate premature, idealised, and delayed take-off. (\cref{fig:baton}). These take-off conditions are determined by the ratio of gravitational to centripetal forces. 

Inspired by \citep{Baton_Roberts_1979} the rotational jumping model is a massless rod with a length, $d$, and a point mass, $m_B$, at one end (\cref{fig:baton}a). The other end of the rod is connected to the ground via a revolute joint and a massless linear (Hookean) rotational spring with a stiffness, $k_r$ and a natural spring angle, $\theta_{ini}$. There is no unsprung mass or unsprung inertia in this example model. The kinematics of this model is captured by the rotational displacement of the body, $\theta$, measured from the ground plane. $y$ is the vertical displacement of the body measured from the ground and is collinear but opposite to gravity and is derived as $y=d\sin{\theta}$; $x$ is the horizontal displacement of the body measured from the ground contact point F and is derived as $x=d\cos{\theta}$.

The jumping process begins at the standing posture where the rotational spring has already been compressed by the weight of the point mass, $m_Bg$ (\cref{fig:baton}c). A spring charging torque, $\tau_c$, is applied to the point mass to charge the rotational spring. The compression is stopped when the spring is fully charged at $\theta=0\degree$, where the rod is collinear with the $x$-axis; this state is defined as a \textit{charged}. The total elastic energy stored in the spring is $\frac{1}{2}k_r(\theta_{ini})^2$. Once the spring-charging torque is removed, the rod accelerates in $+\theta$ direction. The equation of motion of the system is,
\begin{equation}
    m_Bd\ddot{\theta}=\frac{\tau_k}{d}-m_Bg\cos{\theta},\label{eq_baton_EOM}
\end{equation}
where $\tau_k$ is the spring restoring torque and $\tau_k=k_r(\theta_{ini}-\theta)$. Note that as $\theta$ tends to zero, \cref{eq_baton_EOM} becomes equivalent to \cref{eq_pris_EOM}. The tension, $T$, exerted on the rod is the sum of the gravitational load and the centripetal force, $F_{cen}$,  as follows,
\begin{equation}
    T=m_Bg\sin{\theta}-F_{cen}=m_Bg\sin{\theta}-m_Bd\dot{\theta}^2.\label{eq_baton_Tension}
\end{equation}

The vertical ground reaction force acting on the bottom end of the rod, $F_R$, is the sum of the vertical component of the spring restoring torque, $\frac{\tau_k}{d}\cos{\theta}$, and the vertical component of the rod tension, $T\sin{\theta}$, which is expressed as the weight component, $m_Bg\sin^2{\theta}$, minus the vertical component of the centripetal force, $m_Bd\dot{\theta}^2\sin{\theta}$, 
\begin{equation}
    F_R=\frac{\tau_k}{d}\cos{\theta}+T\sin{\theta}=\frac{\tau_k}{d}\cos{\theta}+m_Bg\sin^2{\theta}-m_Bd\dot{\theta}^2\sin{\theta}.\label{eq_baton_FR}
\end{equation}

Take-off occurs when $F_R=0$ and can be realised if all of the three force components equal zero, or if they are nonzero and cancel each other.  

\cref{fig:baton}d-i shows examples of the jumping dynamics of the rotational model using different spring stiffness values to cause different take-off conditions. The normalised spring stiffness is the rotational spring stiffness normalised by the system characteristic length and weight as $k_r/mgd$. With a normalised rotational spring stiffness of 4.5, the model undergoes a delayed take-off (\cref{fig:baton}d,g); a normalised value of 5.5 results in an idealised take-off (\cref{fig:baton}e,h), while a value of 10 leads to premature take-off (\cref{fig:baton}f,i). These three distinct dynamic behaviours can occur for any jumping system comprised of rotational links. 

These take-off conditions can be explained by examining the three force components acting on the ground (\cref{fig:baton}g-i). Throughout the acceleration phase, the spring restoring torque decreases and the rod angle increases, which both cause the first term on the right-hand side of \cref{eq_baton_FR}, $\frac{\tau_k}{d}\cos{\theta}$, to decrease. But simultaneously, the spring torque accelerates the point mass and, in combination with the rising rod angle, causes an increase in the gravitational force component, $m_Bg\sin^2{\theta}$, and the centripetal force component, $m_Bd\dot{\theta}^2\sin{\theta}$, which are denoted as the second and third terms on the right-hand side of \cref{eq_baton_FR}.

The restoring torque is proportional to the spring stiffness, whereas the centripetal force is proportional to the angular velocity squared (\cref{eq_baton_FR}). As the spring stiffness increases the restoring torque component increases, but the angular velocity and centripetal force component increase by a greater amount (moving from (d) to (e) to (f) and from (g) to (h) to (i) in \cref{fig:baton}). Therefore, as the spring stiffness increases, the centripetal force term cancels out the torque term at a lower displacement. This causes the take-off to occur at a lower displacement. The premature take-off is evident from the reduction in the take-off angle compared to the natural spring angle. This dynamic event is undesirable in a spring-driven jumping system as it decreases the elastic-kinetic energy efficiency. However, it is noteworthy that even though the premature take-off is exacerbated with the increased normalised spring stiffness, the increased spring restoring torque increases the angular take-off velocity of the rotational system.  This implies that despite the decrease in elastic-kinetic energy efficiency due to increased spring torque, the system can achieve a greater jump height or distance as the take-off velocity increases.

In practice, a physical multibody system may involve multiple rotational masses, but the fundamental principle remains consistent. To attain an idealised take-off, the gravitational force component must be equal to the centripetal force components at the moment when the ground reaction force diminishes to zero. \cref{sec:results_efficiency} will show that the premature take-off is exacerbated not only by the increasing spring torque but also by a higher proportion of rotational masses within the multibody system.

\section{Jumping dynamics of a multibody system}
\label{sec:multibody}
The previous section studied the jumping dynamics of the abstracted prismatic and rotational models powered by springs. However, physical jumping systems consist of linkages with multiple segments and joints, leading to more complex dynamics. This section will develop a multibody dynamic model to explore the coupled dynamic effects and identify the inefficient energy source. A case study will be conducted using a rhomboidal linkage with rotational springs (rotational spring-linkage), which serves as a representative example of a generalised spring-driven jumping system. A physical experiment will then be used to validate the dynamic model.  \cref{sec:results_exp,sec:results_efficiency} will present the findings with solutions to increase the elastic-kinetic energy conversion efficiency in practical spring-driven jumping robots.

\subsection{Model kinematics}
\label{sec:multibody_kinematics}

\begin{figure}[t]
\centering
\includegraphics[scale=0.99]{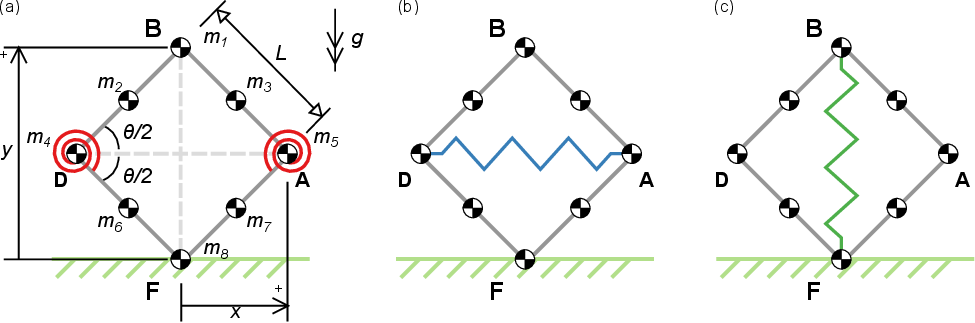}
\caption{Examples of multibody systems formed by the rhomboidal linkage with (a) rotational springs, (b) horizontally placed translational spring and (c) vertically placed translational spring \citep{LO_Statics_2023}.}
\label{fig:rhom}
\end{figure}

A multibody representation of the rotational spring-linkage system is defined using four rigid segments connected by revolute joints in \cref{fig:rhom}. The rhomboidal linkage is one of the most common topologies in jumping robots and has a single translational degree of freedom (DoF) \citep{JReview_Zhang_2017,JReview_Zhang_2020,JReview_Ribak_2020,JReview_Mo_2020}.  This mechanism is chosen here as it is simple enough to provide interpretable results in multibody dynamics, while also maintaining adequate detail to represent existing jumping robot designs. Point masses $m_1$, $m_4$, $m_5$ and $m_8$ are positioned at the revolute joints and represent the mass of the mechanical joints A, B, D and F including the mass of the rotational springs. Joint B represents the body, joint F represents the ground contact point of the system (e.g. foot), and joints A and D are the knee joints. The angle between the segments BA and FA, $\theta$, is defined as the \textit{knee angle} and is equal to the angle of the rotational spring when this spring type is used. The rigid segments are modelled as thin uniform beams of mass $m_2$, $m_3$, $m_6$ and $m_7$ have equal length, $L$; note that $L=d/2$, where $d$ is the characteristic length of the system. The centre of mass of each beam is at the geometric centre, and the moment of inertia of each beam about its centre of mass is given by $I_i=m_iL^2/12$. It is important to include both the joint masses and the segment rotational inertias within the model as these will both be shown to contribute to the take-off dynamics. $y$ is the vertical displacement of joint B measured from the ground plane and is aligned with, but opposite to, the local acceleration due to gravity; $x$ is the horizontal displacement measured from the initial position of joints B and F, and is perpendicular to $y$. The translational position, velocity and acceleration of joint B are given by, 
\begin{equation}
   x=L\cos\frac{\theta}{2},\dot{x}=-\frac{L\dot{\theta}}{2}\sin\frac{\theta}{2},\ddot{x}=-\frac{L}{2}(\ddot{\theta}sin\frac{\theta}{2}+\frac{\dot{\theta}^2}{2}\cos\frac{\theta}{2}); \label{eq_rhom_x}
\end{equation}
\begin{equation}
   y=2L\sin\frac{\theta}{2},\dot{y}=L\dot{\theta}\cos\frac{\theta}{2},\ddot{y}=L(\ddot{\theta}\cos\frac{\theta}{2}-\frac{\dot{\theta}^2}{2}\sin\frac{\theta}{2}). \label{eq_rhom_y}
\end{equation}

\subsection{Jumping dynamics}
\label{sec:multibody_dynamics}

The dynamics of the system are analysed by the Lagrangian method given by,
\begin{equation}
   \mathcal{L}=KE-PE,\label{eq_rhom_Lagran}
\end{equation}
\begin{equation}
    \frac{\partial}{\partial t}(\frac{\mathcal{L}}{\partial \dot{\theta}})=\frac{\partial \mathcal{L}}{\partial \theta},\label{eq_rhom_Lagran_2}
\end{equation}
where $\mathcal{L}$ is the Lagrangian. $KE$ is the total kinetic energy, which is the sum of the horizontal kinetic energy, $KE_x$, vertical kinetic energy $KE_y$, and rotational kinetic energy $KE_r$,
\begin{equation}
    KE=KE_x+KE_y+KE_r=\sum_{i=1}^{8} \frac{1}{2}m_i(\dot{x}_i)^2+\sum_{i=1}^{8} \frac{1}{2}m_i(\dot{y}_i)^2+\sum_{i=1}^{8} \frac{1}{2}I_i(\dot{\theta}_i)^2, \label{eq_rhom_KE_T}
\end{equation}
where $m_i$ is the mass of the $i$th component (joint or segment), $x_i$ and $y_i$ are the positions of the respective centres of mass. Note that in this class of 1-DoF system the centre of mass has zero horizontal and rotational kinetic energy. The total vertical kinetic energy is the sum of the vertical kinetic energy of the system centre of mass, $KE_{y,CG}$, and the vertical kinetic energy of the components relative to the centre of mass, $KE_{y,relative}$:
\begin{equation}
    KE_y=KE_{y,CG}+KE_{y,relative}=\frac{1}{2}m_T(\dot{y}_{CG})^2+\sum_{i=1}^{8} \frac{1}{2}m_i(\dot{y}_i-\dot{y}_{CG})^2, \label{eq_rhom_KEy}
\end{equation}
where the summation of the eight components includes the four revolute joints and the four segments. $PE$ is the total potential energy, including the gravitational potential energy, $GPE$, and the elastic potential energy stored in the rotational springs, $EPE$:
\begin{equation}
    PE=GPE+EPE=\sum_{i=1}^{8} m_ig\dot{y}_i+k_r(\theta_{ini}-\theta)^2, \label{eq_rhom_PE_T}
\end{equation}
where $g$ is the gravitational acceleration, $k_r$ is the rotational spring stiffness and $\theta_{ini}$ is the natural angle of the rotational spring. By substituting \cref{eq_rhom_KE_T,eq_rhom_PE_T} into \cref{eq_rhom_Lagran}, the Lagrangian can be rewritten as,
\begin{equation}
   \mathcal{L}=\frac{L^2\dot{\theta}^2}{32}m_A\sin^2\frac{\theta}{2}+\frac{L^2\dot{\theta}^2}{32}m_B\cos^2\frac{\theta}{2}+\frac{L^2\dot{\theta}^2}{96}m_C-m_D\frac{gL}{2}\sin\frac{\theta}{2}-k_r(\theta_{ini}-\theta)^2, \label{eq_rhom_Lagran_3}
\end{equation}
where $m_A=\sum_{i=2}^7 m_i+3\sum_{i=4}^5 m_i$, $m_B=\sum_{i=1}^7 m_i+3\sum_{i=1}^5 m_i+5\sum_{i=1}^3 m_i+7m_1$, $m_C=\sum_{i=2}^7 m_i-\sum_{i=4}^5 m_i$, $m_D=\sum_{i=1}^7 m_i+\sum_{i=1}^5 m_i+\sum_{i=1}^3 m_i+m_1$. By substituting \cref{eq_rhom_Lagran_3} into \cref{eq_rhom_Lagran_2}, the angular acceleration of the knee can be expressed as,
\begin{equation}
   \ddot{\theta}=\frac{2k_r(\theta_{ini}-\theta)-\frac{L^2\dot{\theta}^2}{64}\sin\theta(m_A-m_B)-\frac{m_DLg}{4}\cos\frac{\theta}{2}}{\frac{L^2}{16}(m_A\sin^2\frac{\theta}{2}+m_B\cos^2\frac{\theta}{2})+\frac{L^2}{48}m_C}. \label{eq_rhom_Lagran_ar}
\end{equation}

The vertical ground reaction force, $F_R$, acting on joint F is,
\begin{equation}
   F_R=m_Tg+m_T\ddot{y}_{CG}=\sum_{i=1}^8 m_i(g+\ddot{y}_i), \label{eq_rhom_FR}
\end{equation}
where $m_T$ is the total mass of the system and $\ddot{y}_{CG}$ is the vertical acceleration of the centre of mass of the entire system.

\subsection{Take-offs and jump height}
\label{sec:multibody_takeoff}

The jumping spring-linkage takes off when the ground reaction force equals zero ($F_R=0$) and the acceleration of the centre of mass is equal to $-g$. This condition can be expressed by substituting \cref{eq_rhom_y,eq_rhom_Lagran_ar} into \cref{eq_rhom_FR}; the stiffness of the rotational spring is obtained by the given peak spring-charging force, $F_{max}$, natural spring angle and the charged angle, $\theta_{end}$, using Eq. (14) in \citep{LO_Statics_2023}, as follows,
\begin{equation}
   -\frac{4m_Tg}{m_D}=\frac{[\alpha m_Tg\cos\frac{\theta_{end}}{2}(\frac{\theta_{ini}-\theta_{to}}{\theta_{ini}-\theta_{end}})-\frac{m_Dg}{4}\cos\frac{\theta_{to}}{2}]\cos\frac{\theta_{to}}{2}-\frac{L(\dot{\theta}_{to})^2}{96}(3m_A+m_C)\sin\frac{\theta_{to}}{2}}{\frac{1}{16}(m_A\sin^2\frac{\theta_{to}}{2}+m_B\cos^2\frac{\theta_{to}}{2})+\frac{1}{48}m_C}, \label{eq_rhom_takeoffcondition}
\end{equation}
where $\alpha=F_{max}/m_Tg$ is the force-to-weight ratio, $\theta_{to}$ is the take-off spring angle and $\dot{\theta}_{to}$ is the take-off angular velocity. 

Throughout the acceleration phase the total energy of the system remains constant and equal to the initial elastic potential energy stored in the spring so that, $KE+PE=EPE_{ini}=k_r(\theta_{ini})^2$. At take-off, the residual elastic potential energy and the kinetic energies of the components are generally regarded as the sources of inefficiency as they represent the portion of stored elastic potential energy that is not converted to gravitational potential energy inflight. The vertical take-off velocity of the centre of mass is,
\begin{equation}
   \dot{y}_{CG,to}=\frac{\sum_{i=1}^{8} m_i\dot{y}_{i,to}}{m_T}=\frac{m_D}{4m_T}\dot{y}, \label{eq_rhom_vyCGto}
\end{equation}
where $\dot{y}_{i,to}$ is the vertical take-off velocity of each mass. The jump height of this system, $h$, measures the maximum vertical displacement of the centre of mass from its displacement at take-off. By conservation of energy, it is derived from the take-off velocity of the centre mass, $\dot{y}_{CG,to}$, as,
\begin{equation}
   h=\frac{1}{2g}(\dot{y}_{CG,to})^2. \label{eq_rhom_h}
\end{equation}
This work compares the jump height of different scales of systems by normalised jump height, $\tilde{h}$, which is defined as the ratio of the jump height to the characteristic length of the system as,
\begin{equation}
   \tilde{h}=h/d. \label{eq_rhom_nh}
\end{equation}

As an example in jumping robots, the hybrid spring-linkage jumper \citep{NatureJ_Hawkes_2022} uses a 0.3m nonlinear spring-linkage to jump over 33m high, which is equivalent to a normalised jump height of 110. 

For reference, the theoretical normalised jump height upper bound is set by an idealised system (e.g. the prismatic model in \cref{sec:prismatic} without unsprung mass) driven by an ideal constant force spring. Here, the force of the ideal spring would be independent of the spring displacement. During the spring-charging phase, the ideal spring would be charged by a constant force equal to the sum of the peak motor force and system weight, $F_{max}+m_Tg$. Thus, it conceptually stores the maximum amount of elastic energy for a given mass, motor force and characteristic length. The jump height of this ideal system is derived from the stored elastic energy by assuming the conservation of energy as, $m_Tg(h+d)=(F_{max}+m_Tg)d$, so that, $h=F_{max}d/m_T$. The normalised jump height of the ideal spring-driven system is equal to the force-to-weight ratio as $\tilde{h}=F_{max}/m_Tg=\alpha$. In contrast, the prismatic linear spring-driven model with zero unsprung mass (\cref{sec:prismatic}) can only store 50\% energy of the ideal system so that $h=\frac{d}{2}(\frac{F_{max}}{m_Tg}-1)$ and results in $\tilde{h}=\frac{1}{2}(\alpha-1)$. 

\section{Experimental model}
\label{sec:exp_model}
\begin{figure}[t]
\centering
\includegraphics[scale=1]{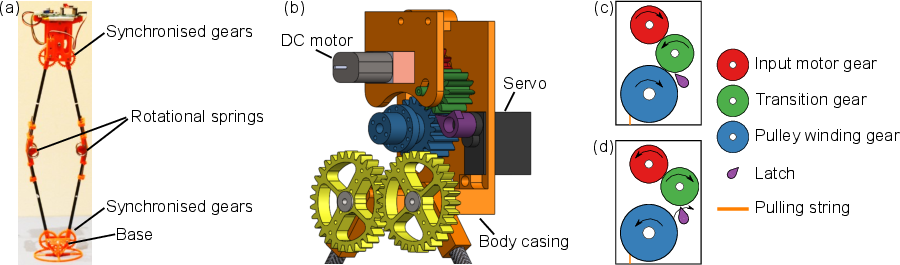}
\caption{(a) The experimental model and (b) the CAD model of the main body. Working schematic of the reduction gearbox with the active latch in (c) the spring-charging and (d) the acceleration phases.}
\label{fig:exp_model}
\end{figure}

An experiment is conducted to validate the acceleration dynamics predicted by the theoretical models. The experiment includes a self-contained jumping system with the spring-linkage arrangement used in the previous section. The system has onboard power, a control system that initiates spring charging, and an active latch to start the acceleration phase (\cref{fig:exp_model}a). The system can undertake multiple charging and jumping cycles.

The jumping system has a mass of 206g and has a height of 40cm. \cref{tab:mass_budget} lists the mass breakdown. The single degree of freedom is realised by the pairs of synchronised spur gears at joints B and F, inspired by \citep{JPL_Hale_2000}, as shown in \cref{fig:exp_model}(a).

The motor module comprises a three-stage reduction gearbox (1.5:1), a DC motor (6V, Pololu 1000:1) and a servo (Turnigy D56MG), as shown in \cref{fig:exp_model}(b). The gearbox increases the torque output from the DC motor shaft, and connects the motor to a pully to wind a tension string when charging the robot. The servo serves as a latch to actively release the charged springs. The maximum output torque and the string tensile force of this module are 0.37Nm and 60N, respectively.

During the spring charging phase, the motor gear contacts the transmission gear which is engaged with the pully winding gear to wind the string, as shown in \cref{fig:exp_model}(c). To initiate the jump, the latch triggered by the servo displaces the transmission gear to disconnect the DC motor from the drivetrain, as shown in \cref{fig:exp_model}(d). Without the spring holding force, the embedded springs can recoil freely and release the stored elastic energy, causing the leg to extend. Low friction bearings MF52ZZ are installed in each revolute joint and gearbox to minimise friction during the acceleration phase.

The system employs an Adafruit ESP32 microprocessor to remotely communicate and control the system with Bluetooth via a smartphone. The power system consists of a 3.7V LiPo battery and a boost-regulator MT3608. The battery is selected to match its available discharging current with the maximum current draw from the DC motor. The regulator boosts the voltage for the 6V DC motor and the servo.

The experimental model uses two pairs of rotational springs attached to its knee joints. Each rotational spring pair has a stiffness of 0.7Nmrad$^{-1}$ and a natural spring angle of 178$\degree$. An experiment was conducted to validate the spring properties, as shown in \citep{LO_Statics_2023}. The initial knee angle of the robot is 178$\degree$, aligning with the natural spring angle; there was no observable deflection caused by the system weight. These springs are selected so that only half of the available motor force from the motor module is needed to compress the system, providing a system safety factor of 2. Note that the required motor force is derived by applying the quasi-static equation from \citep{LO_Statics_2023} with the given natural spring angle and leg length of the robot.

The prototype is initially placed on a horizontal surface in the standing posture. Both the spring charging and the acceleration phases are triggered remotely. The take-off kinematics were captured using a high-speed camera Phantom V310 at a frame rate of 1000 FPS. The inflight trajectory is recorded using a GoPro Hero 6 at 120 FPS. The jumping kinematic data were obtained from the video using the automated pixel tracking software, \textit{Tracker}, to capture the positions of the joints A,B,D and F and the centres of each link at time intervals of 1ms. 

\section{Numerical model}
\label{sec:num_model}
The dynamics model in \cref{sec:results_exp} is used to predict the acceleration phase of the experimental model up to the time of take-off. It is also used in \cref{sec:results_efficiency,sec:results_nh} to simulate the acceleration phase of the conceptual models based on the experimental model that cannot be tested by experiment. The acceleration phase is modelled through a numerical solution to \cref{eq_rhom_Lagran_ar} using Matlab’s ODE45 solver with a maximum variable time step size of 10$^{-4}$s. The model parameters are $L=0.15$m, $k_r=0.7$Nmrad$^{-1}$, $\theta_{ini}=178\degree$ and $\theta_{end}=25\degree$. The mass distribution of the dynamic model is shown in \cref{tab:mass_budget}.  

A second numerical multibody dynamics model is developed using the Matlab Simulink (Simscape Multibody toolbox library) to verify the results from the ODE45 solution. The toolbox simulates the system jumping and inflight dynamics under gravitational forces, inertial forces and moments and the spring driving force and moments. The ground is as being fixed to the world frame. The contact force between the foot and the ground is modelled as a linear spring with a stiffness of 10$^8$ Nm$^{-1}$. The geometry of the multibody model is equal to that of the dynamic model in \cref{tab:mass_budget}. The model has a single translational degree of freedom aligned with the gravitational vector and is realised by a prismatic joint connecting the body of the model to the ground. The simulation is computed using a fixed timestep solver (ODE14x) with a time step of 10$^{-5}$s; further reduction in timestep yielded a change of less than 0.1\% in all results.  

\begin{table}[t]
\begin{center}
\begin{tabular}{|c|c|c|c|}
\hline
\textbf{Dynamic model} & \textbf{Physical model}& \textbf{Mass (g)}& \textbf{Mass fraction}\\
\hline
\multirow{7}{*}{$m_1$}& Motor&11.5&\multirow{7}{*}{56\%}\\
\cline{2-3}
          & Gears                      & 8 &\\
\cline{2-3}
          & Latch                      & 10 &\\
\cline{2-3}
          & Body casing                & 31 &\\
\cline{2-3}
          & Control unit               & 32 &\\
\cline{2-3}
          & LiPo battery               & 18 &\\
\cline{2-3}
          & Synchronised gears (upper) & 5  &\\
\hline
$m_2,m_3$ &Upper legs                  &8.1 &4\%\\
\hline
\multirow{2}{*}{$m_4, m_5$}& Rotational springs & 48 & \multirow{2}{*}{31\%}\\
\cline{2-3}
          & Joints                     & 15.4  &\\
\hline
$m_6,m_7$ &Lower legs                  &8.1 &4\%\\
\hline
\multirow{2}{*}{$m_8$}& Synchronised gears (lower) & 5 & \multirow{2}{*}{5\%}\\
\cline{2-3}
          & Base                       &5.5 &\\
\hline
 & \textbf{Total} &205.6 & 100\%\\
\hline
\end{tabular}
\caption{The mass budget of the experimental model.}
\label{tab:mass_budget}
\end{center}
\end{table}

\begin{figure}[b]
\centering
\includegraphics[scale=1]{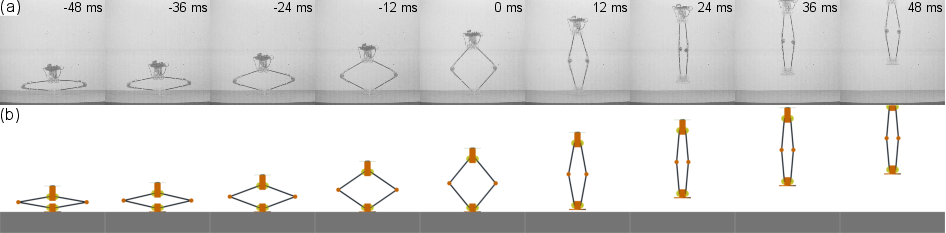}
\caption{(a) Take-off sequence of the prototype and (b) the Simscape model, starting from the charged posture and then accelerating to the take-off position before becoming airborne. The simulation time is the same as the experiment.}
\label{fig:phantomCAD}
\end{figure}

\section{Result and discussion}
\label{sec:results}
This section will first compare the results of the jump dynamics of the experiment and, the mathematical model given in \cref{sec:multibody} and the numerical model given in \cref{sec:num_model}. It will then discuss the influence of the motor force and mass distribution on the take-off dynamics and the elastic-kinetic energy conversion efficiency of the multibody. 

\subsection{Experiment and simulation result of the experimental model}
\label{sec:results_exp}
\cref{fig:phantomCAD}(a) illustrates the jumping sequence of the experimental model. The jump starts from the charged position, followed by the acceleration phase, eventually reaching the take-off state (at $t=0$ms) and becoming airborne. \cref{fig:phantomCAD}(b) is a digital visualisation of the experiment using the Simulink model mentioned in \cref{sec:num_model}; it gives identical results to the numerical model from \cref{sec:multibody}, but is shown here as it also provides an illustration of the airbourne phase.

\begin{figure}[ht]
\centering
\includegraphics[scale=1]{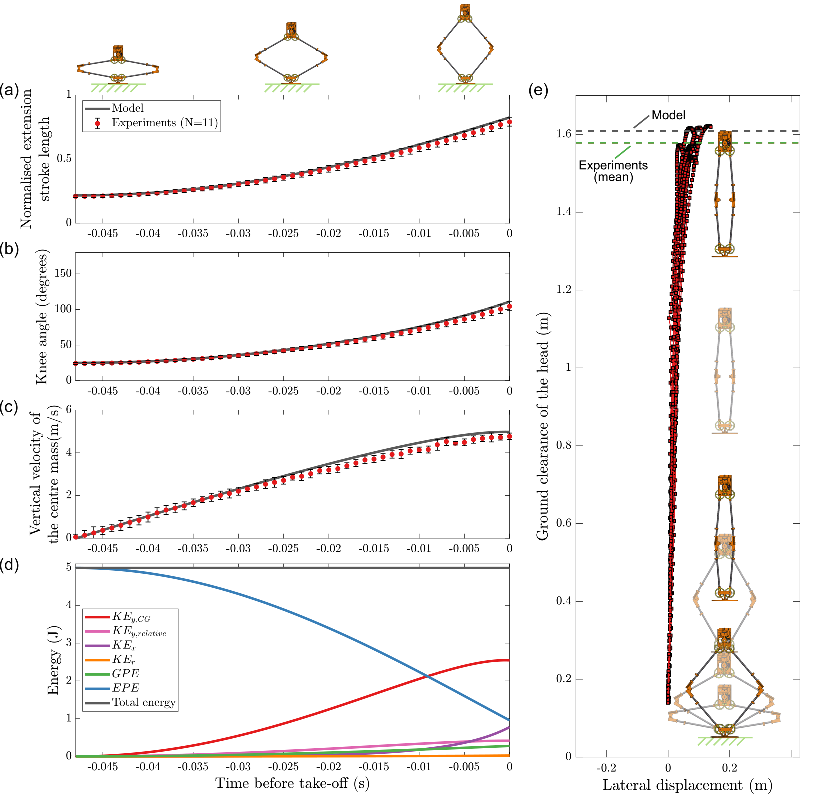}
\caption{Experimental and theoretical data of the system during a jump. (a)-(c) depict the kinematics up to the point of take-off measured experimentally and predicted by the numerical model: (a) Extension stroke length normalised by the length of the robot, (b) the knee angle and (c) the velocity of the centre of mass. (d) Energy components of the system are obtained from the numerical model. (e) Experimental measurement of the jumping trajectory during take-off and flight, measured as the ground clearance of the top of the body; the mean value of the trajectory maximum ground clearance from each experiment is shown as the green dashed line, and the numerically simulated maximum ground clearance is shown as the black dashed line.}
\label{fig:proto_exp}
\end{figure}

\cref{fig:proto_exp}(a)-(c) shows the experimental results from the 11 trials with the numerical result shown as the solid dark line. The initiation of the acceleration phase is defined here as when the centre of mass velocity is greater than 0.5ms$^{-1}$. From the charged to take-off states the acceleration phase lasts around 0.05s. The energy of the system is calculated from the numerical model, as shown in \cref{fig:proto_exp}(d). \cref{fig:proto_exp}(e) shows the jumping trajectory of the prototype represented by its body displacement from the ground. The dashed grey line marks the prediction from the numerical model; the dashed green line marks the mean peak vertical displacement in the experiments.   

The numerical model closely predicts the trends seen in the experiments in \cref{fig:proto_exp}(a)-(c). The experimental model was compressed to a knee angle of 25$\pm$1$\degree$, equivalent to a normalised stroke length of 0.2$\pm$0.01. Around 5J of elastic potential energy was stored in the pair of rotational springs, as shown in \cref{fig:proto_exp}(d). The experimental model accelerated at over $10g$, and took off prematurely with a knee angle of 104$\pm$4$\degree$; note that an idealised system would have a take-off knee angle equal the natural angle of the rotational spring of 178$\degree$. As explained in \cref{sec:baton}, the premature take-off was due to the centripetal force on the segments and knee joints. The vertical take-off velocity of the centre of mass is 4.8$\pm$0.2ms$^{-1}$, which is 4.1\% lower than the prediction of the numerical model. The elastic-kinetic energy conversion efficiency of the experiment is 0.47$\pm$0.03 (solid red line in \cref{fig:proto_exp}d), 8.1\% lower than the model. The difference between the experiment and the predictions is believed to be mainly attributed to the factors that are not captured in the model, including the air resistance, the structural friction and the damping of the physical substrate. At the apex, the top of the body was measured as being 1.58$\pm$0.03m from the ground, 1.9\% lower than the prediction of the model.

The total system energy is equal to the initial elastic potential energy charged in the rotational springs and remains unchanged throughout the acceleration phase (\cref{fig:proto_exp}d). When the latch releases the springs, the stored elastic potential energy is converted into gravitational potential energy and kinetic energy as the linkage begins to extend and accelerate. The numerical model gives a breakdown of the transfer of energy into various sources, providing insight into sources of inefficiency (\cref{fig:proto_exp}d). By the time of take-off 6\% of the stored elastic energy has been converted to gravitational potential energy and 51\% has been converted to the vertical kinetic energy of the centre of mass. The latter measure is typically the desirable quantity for a jumping robot as it defines the eventual jump height. The remaining 43\% of initially stored elastic energy is not converted to useful work done on the system: 19\% of the elastic potential energy remains in the spring as residual energy at the point of take-off, hence we define this as a premature take-off. Around 24\% of the elastic energy is used to accelerate the individual segments relative to the overall system centre of mass: 16\% is transferred to the horizontal motion of the segments, 8\% to vertical motion, and less than 1\% to rotational motion. In flight these energy components are observable as oscillations of the segments. In the experiments the oscillations decay due to structural and aerodynamic damping, and all components are seen to travel at approximately the same velocity relative to the Earth immediately prior to landing.

In the second experiment the effect of mass distribution on jump performance was explored by adding extra masses to the experimental model. 200g additional mass was distributed between the body and foot of the robot. \cref{fig:proto_payload}a-b show the variation in elastic-kinetic energy conversion efficiency and normalised jump height with body mass fraction for the physical experiments (red dots) and numerical models of the experiments (solid red line). As the body mass fraction increases, the energy efficiency and normalised jump height increase. For the given experiment the maximum attainable energy efficiency is up to 0.6 with a normalised jump height of 2.4 at around 78\% body mass fraction, which is limited by the inherent weight of the segments and foot in the current design. The cause of this performance improvement will be explained in \cref{sec:results_efficiency} using the numerical model introduced in \cref{sec:multibody}. 

\subsection{The influence of the mass distribution and force-to-weight ratio on the energy efficiency}
\label{sec:results_efficiency}

\begin{figure}[t]
\centering
\includegraphics[scale=1]{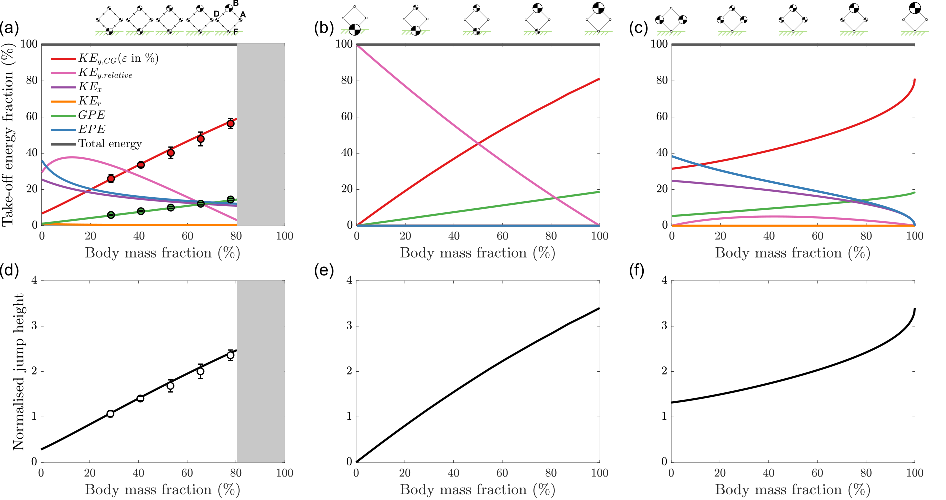}
\caption{The (a) take-off energy fraction and (b) the normalised jump height of the experimental and numerical results of simulated jumps of the modified system with the additional 200g payload. (c-d) Numerical simulations of systems made of a massless linkage with concentrated mass at body and foot. (e-f) Numerical simulations of the system with concentrated mass at the body and knees. In (a-b), the greyed area is the unattainable region where the body mass fraction cannot be further increased due to the inherent weight of the segments and foot in the experimental model. For (c-f), the massless linkage system cannot be tested experimentally and the results are gained from the numerical model. Note that the take-off energy fraction is the energy components of the system at take-off normalised by the initial elastic energy stored in the rotational springs of the system; the body mass fraction is the proportion of the body mass to the total system mass. The jump height is the vertical displacement of the system centre of mass from take-off to apex calculated by $h=\frac{1}{2g}(\dot{y}_{CG,to})^2$ and is normalised by the characteristic length, $d$.}
\label{fig:proto_payload}
\end{figure}

The numerical model in \cref{sec:multibody} can be used to assign point masses at locations in the system, which allows specific dynamic effects to be controlled to explore their influence on the system performance.  Two theoretical models of a massless linkage system were developed with concentrated masses at the body and foot joints (\textit{body – foot} in \cref{fig:proto_payload}b-c) and at body and knee joints (\textit{body – knees} in \cref{fig:proto_payload}c-d). The body mass fraction was adjusted by varying the mass between the two joints in each model. These conceptual systems have the same dimensions and total mass as the modified experimental model including the additional masses. 

As the mass is distributed higher up the model the elastic-kinetic energy conversion efficiency, $\varepsilon$, increases (\cref{fig:proto_payload}). With all mass located at the body, the body mass fraction is 100\%, the energy efficiency rises to 0.8. Here, all of the spring energy is converted to gravitational potential energy and kinetic energy of the system centre of mass prior to take-off. This is the maximum theoretically achievable elastic-kinetic energy efficiency and normalised jump height for this force-to-weight ratio and characteristic length. But note that the elastic-kinetic energy efficiency is less than 1 due to the elastic energy converted to gravitational potential energy before take-off. Nevertheless, it could be increased further by increasing the system force-to-weight ratio or decreasing the system characteristic length to reduce the gravitational potential energy at take-off. 

The body-foot model (\cref{fig:proto_payload}c-d)  is useful as it removes centripetal forces, horizontal inertial forces, and rotational inertia from the system. This simplifies the energy breakdown in \cref{fig:proto_payload}a. In the body-foot model, the most extreme case is when the body mass fraction equals 0, so the system centre of mass is located at the unsprung static foot. The spring restoring force does no work on the centre of mass as the massless leg extends, so it never takes off. As the body mass fraction increases the elastic-kinetic energy conversion efficiency and normalised jump height increase. This is because the centre of mass is shifted towards the body, and the spring can now do work on the centre of mass. The remaining energy is transferred to the relative motion of the body and foot with respect to the centre of mass and the residual elastic potential energy, as shown in \cref{fig:proto_payload}c. Like the prismatic model in \cref{sec:prismatic}, premature take-off does not occur as the segments and knee joints are massless and have no centripetal force acting on them. Instead, delayed take-off occurs due to the unsprung foot mass. This undesirable dynamic phenomenon could be improved by increasing the body mass fraction. With a higher sprung mass proportion, a lower spring force is needed to pull the static unsprung foot off the ground at take-off. This results in lower residual elastic energy in the spring at take-off, and a higher elastic-kinetic energy efficiency and normalised jump height. 

The body-knees model (\cref{fig:proto_payload}e-f) is used to introduce centripetal effects into the model while removing the unsprung foot mass. In the body-knees models, the presence of the knee mass prevents an idealised take-off. As the leg extends during the acceleration phase, the centripetal force on the knee masses has a vertical component acting on the ground that opposes the system weight and the vertical component of the spring restoring force. As the spring discharges and the rotational inertia accelerates, the centripetal force increases with the angular velocity squared. When the vertical component of the centripetal force cancels out the vertical component of the spring restoring force and weight, the ground reaction force vanishes and the model takes off prematurely. The centripetal force on the knee is the greatest for systems with a body mass fraction equal to zero. In this extreme case, the model takes off the most prematurely with the greatest proportion of residual elastic energy. Increasing the body mass fraction reduces the relative mass and the centripetal force on the knee, and the model has a longer acceleration time to allow the spring to do more work on the centre of mass. Therefore, the vertical velocity of the centre of mass increases with the body mass fraction, and so do the elastic-kinetic energy conversion efficiency and jump height. 
By moving the mass towards the body, both conceptual models show a significant improvement in their jumping performance. These numerical analyses identify the mass distribution of the system as one of the keys to determining the take-off condition and jumping dynamics. 

\begin{figure}[t]
\centering
\includegraphics[scale=1]{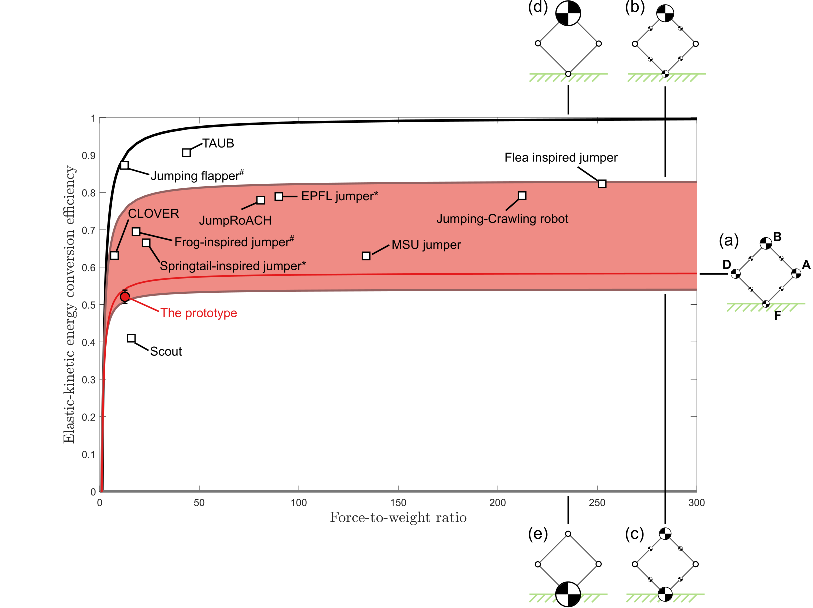}
\caption{The elastic-kinetic energy conversion efficiency variation with the force-to-weight ratio for the size of the experimental model in the present work. The lines represent different mass distributions. (a) (solid red line) mass distribution of the experimental model in this work, with a maximum energy efficiency of around 0.58. The red coloured region shows the predicted energy efficiency if the mass of the knee joints and the rotational springs in the robot were repositioned to (b) the body (joint B) or (c) the foot (joint F). The theoretical bounds on the elastic-kinetic energy efficiency for this size of the system are set by two conceptional models: (d) a  system where the mass is concentrated at the body; (e) a  system with all mass concentrated at the foot, which can never take-off, regardless of the force-to-weight ratio. Experimental measurements from previous jumping robots are taken from \citep{Jumproach_Jung_2016,FleaJ_Koh_2013,ScoutJ_Zhao_2009,FlapJ_Truong_2019,Jumproach_Jung_2019,MSU_Zhao_2013,TAUB_Zaitsev_2015,Springtail_Ma_2021,EPFLJ_Kovac_2008}. *The peak force is approximated by its stored elastic potential energy and its characteristic length of the robotic system by the formula, $F_{max}=\frac{2PE_k}{d}$. \#The peak force is considered to be equal to the peak ground reaction force.}
\label{fig:proto_eff}
\end{figure}

Apart from the mass distribution, another intuitive approach to increasing the elastic-kinetic energy conversion efficiency might be to increase the elastic potential energy stored within the spring. This can be physically achieved using a higher stiffness spring and a higher force motor, which increases the force-to-weight ratio of the system. \cref{fig:proto_eff} illustrates the effects of varying force-to-weight ratio and mass distribution on the experimental model, with other previous jumping robots included for reference. Neglecting the increase in spring and motor mass that would be required to realise this, our numerical model predicts that the elastic-kinetic energy conversion efficiency of the experimental model can only be increased from its current value of around 0.5 to a maximum of around 0.58 (\cref{fig:proto_eff}a). Including the increases in spring and motor mass would result in an even smaller change in the energy efficiency. And this finding holds true for the previous example robots in \cref{fig:proto_eff}. Increasing the force-to-weight ratio beyond the value typically used in practical jumping robots is not an effective approach to increasing the elastic-kinetic energy conversion. 

A more effective approach remains to redistribute mass towards the body of the robot through strategic positioning of mass components or structural optimisation. Using the experimental model as an example, redistributing more mass towards the top of the robot increases the elastic-kinetic energy efficiency (e.g. from (a) to (b) in \cref{fig:proto_eff}) while moving mass towards the foot decreases the efficiency (e.g. from (a) to (c) in \cref{fig:proto_eff}). A practical alternative to efficiently distributing the mass in the experimental model is to reposition the spring joints and their housings to the top joint. This design change has no influence on the charging mechanics \citep{LO_Statics_2023}. For the current force-to-weight ratio, this increases the efficiency from 0.5 to around 0.75, which equates to a 31\% increase in jump height from 1.6m to 2.1m. This elastic-kinetic energy conversion efficiency aligns with that of several previous jumping robots (e.g. EPFL jumper \citep{EPFLJ_Kovac_2008}, flea-inspired jumper \citep{FleaJ_Koh_2013}, as shown in \cref{fig:proto_eff}). Alternatively, a similar effect could be achieved by tapering the leg segment structure toward the unsprung ground contact. Such structural optimisation is case specific and is not in scope for the generalised modelling approach in this present work. However, it poses an interesting route for future studies.

\subsection{Application of the present theory to jumping robot technologies}
\label{sec:results_nh}

\begin{figure}[t]
\centering
\includegraphics[scale=1]{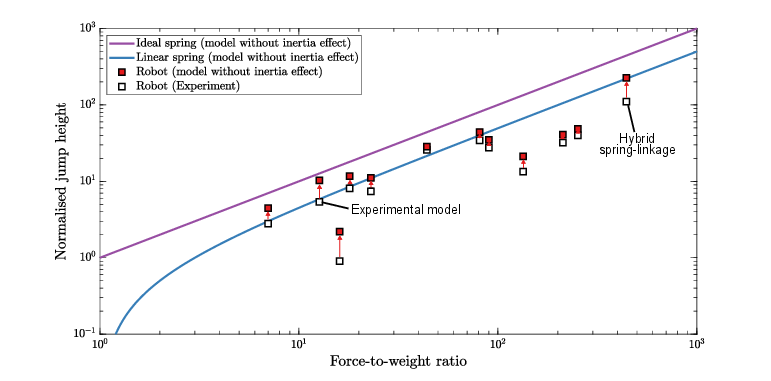}
\caption{The normalised jump height of the experimental model and the previous spring-driven jumping robots \citep{JPL_Hale_2000,Jumproach_Jung_2016,FleaJ_Koh_2013,ScoutJ_Zhao_2009,FlapJ_Truong_2019,Jumproach_Jung_2019,MSU_Zhao_2013,TAUB_Zaitsev_2015,NatureJ_Hawkes_2022,Springtail_Ma_2021,EPFLJ_Kovac_2008}. The white markers indicate the experimental jump height, while the orange markers indicate the predicted jump height when the inertia effect is removed. The jump height of the robots is derived from the experimental measurement of the take-off velocity using the formula, $h=\frac{1}{2g}(\dot{y}_{to})^2$.}
\label{fig:proto_nh}
\end{figure}

The previous section explored the influence of the mass distribution and force-to-weight ratio on the elastic-kinetic energy conversion efficiency. This section aims to apply this knowledge to existing robotic technologies. \cref{fig:proto_nh} shows the jump height of a range of previous jumping robots, with predictions of how their jump heights would increase if the inertial effects of rotational linkage, revolute joints and unsprung foot were removed. Theoretical models presented in \cref{sec:background,sec:multibody} are included for reference. These models represent the theoretical upper limits of jump height for systems driven by a linear spring (\cref{sec:prismatic}) or an ideal constant force spring (\cref{sec:multibody_takeoff}).  

While the experimental dataset in \cref{fig:proto_nh} is not exhaustive, it is suggestive of the fact that as the force-to-weight ratio increases, normalised jump height increases. Previous designs typically achieve lower normalised jump heights than the inertialess linear spring model (e.g. \citep{JPL_Hale_2000,Jumproach_Jung_2016,FleaJ_Koh_2013,ScoutJ_Zhao_2009,FlapJ_Truong_2019,Jumproach_Jung_2019,MSU_Zhao_2013,NatureJ_Hawkes_2022,Springtail_Ma_2021,EPFLJ_Kovac_2008}). The highest jumping robot \citep{NatureJ_Hawkes_2022} reaches around 25\% of the inertialess ideal spring and 50\% of the inertialess linear spring. 

By reallocating all system mass to the top of the structure, the inertial effect can be eliminated. Obviously this is not practically feasible, but it does provide a useful upper limit to bind the performance capabilities of individual designs. For \citep{ScoutJ_Zhao_2009,NatureJ_Hawkes_2022}, removing inertial effects would more than double the jump height. For the other robots shown in \cref{fig:proto_nh}, removing inertial effects would increase the jump height by 11-58\%.

\section{Conclusion}
\label{sec:conclusion}

This paper analysed the jumping dynamics of the abstracted prismatic and rotational spring-driven models and used them as the foundations to study a multibody model of a rhomboidal linkage, as a representation of a physical jumping system. In doing so, it identifies the primary sources of inefficiency as residual elastic potential energy in the springs at the point of take-off, and the motion of the individual segments relative to the centre of mass. The modelling result is experimentally validated by a physical demonstrator, which converts around 50\% of the stored elastic potential energy into vertical kinetic energy of the centre mass. 

From the study of the numerical and experimental models, the amount of residual elastic potential energy in the springs is exacerbated by premature take-off, which is induced by the centripetal forces on the rotational bodies in the linkages. This undesirable dynamic phenomenon could be improved by reducing the overall structural mass of the leg, however this task is non-trivial and would require a case-specific structural redesign. Energy loss to the horizontal and rotational motion will always occur with a rotational linkage; even the highest jumping spring-linkage was estimated to lose 50\% of its stored energy to the inertial effect \citep{NatureJ_Hawkes_2022}.   A solution to this is using an alternative structural configuration without rotational components, such as a prismatic linkage with a linear spring \citep{JPL_Fiorini_1999}. However, these have been shown to offer considerably lower energy storage capacity for a given spring-charging force \citep{LO_Statics_2023} and will suffer from the delayed take-off due to the nonzero unsprung ground interface. So there is a direct tradeoff in mechanism design between topologies that are effective for storing spring energy, and those that are effective for releasing spring energy. Because of this, it can be hypothesized that no ideal jumping system with 100\% mechanical-kinetic efficiency exists, even when frictional effects are neglected. These findings hold true for the system if it were to be scaled in size, but retain the same charging force, mass, and inertia. A practicable design option identified for improving the performance of the jumping robot involves redistributing component mass towards the top of the robot and structural optimisation of the links to taper the mass towards the bottom of the robot.

\bibliographystyle{elsarticle-num} 
\bibliography{Reference_thesis}





\end{document}